\documentclass[12pt,paper]{article}

\usepackage{times}
\usepackage{epsfig}
\usepackage{graphicx}
\usepackage{amsmath}
\usepackage{amssymb}
\usepackage{multirow}

\newcommand{\Tr}{\mathrm{T}}

\newcommand{\PP}{\mathcal{P}}
\newcommand{\NN}{\mathcal{N}}
\newcommand{\TT}{\mathcal{T}}
\newcommand{\XX}{\mathcal{X}}

\newcommand{\mypara}[1]{{\noindent \bf{#1.} }}

\usepackage[pagebackref=true,breaklinks=true,letterpaper=true,colorlinks,bookmarks=false]{hyperref}

\begin{document}

%%%%%%%%% TITLE
%\title{Cross-version metadata mapping in large-scale video collections}
\title{The Video Genome}

\author{Alexander M. Bronstein\\
%Dept. of Computer Science\\
%Technion, Haifa 32000, Israel\\
{\tt\small alex@bbktech.com}
\and
Michael M. Bronstein\\
%Dept. of Computer Science\\
%Technion, Haifa 32000, Israel\\
{\tt\small michael@bbktech.com}
\and
Ron Kimmel\\
{\tt\small ron@bbktech.com}\\
BBK Technologies ltd.\vspace{2mm}\\
Dept. of Computer Science\\
Technion, Haifa 32000, Israel
}

\maketitle
\thispagestyle{empty} % *** Uncomment this line for the final submission

%%%%%%%%% ABSTRACT
\begin{abstract}

Fast evolution of Internet technologies has led to an explosive growth of video data available in the public domain and created unprecedented challenges in the analysis, organization, management, and control of such content. The problems encountered in video analysis such as identifying a video in a large database (e.g. detecting pirated content in YouTube), putting together video fragments, finding similarities and common ancestry between different versions of a video, have analogous counterpart problems in genetic research and analysis of DNA and protein sequences.
In this paper, we exploit the analogy between genetic sequences and videos and propose an approach to video analysis motivated by genomic research. Representing video information as \emph{video DNA} sequences and applying bioinformatic algorithms allows to search, match, and compare videos in  large-scale databases.
We show an application for content-based metadata mapping between versions of annotated video.

\end{abstract}

%%%%%%%%% BODY TEXT
\section{Introduction}

Today, the amount of video content available in the public domain is huge, exceeding millions of hours, and is rapidly growing. Similar growth characterizes video-related \emph{metadata} such as subtitle tracks and user-generated annotations and tags. However, these two types of information belong to two separate and largely unbridged domains. For example, English subtitles available on a DVD version of the \emph{Godfather} movie are hard-wired to the timeline of the DVD video and cannot be used with a different version of the movie, e.g. downloaded from Bittorrent, streamed from YouTube, or broadcast over the air, which has a different timeline.
Similarly, user-generated annotations and comments of a YouTube fragment of the \emph{Godfather} are not accessible to a user watching the movie on DVD.
% (Figure~\ref{fig:eco}, left).
%
%Even if one wished to use the subtitles

%
%\begin{figure*}[t!]
%\begin{minipage}[b]{0.5\linewidth}
%  \centering \includegraphics*[width=0.85\columnwidth]{images/fig1a.eps}
%\end{minipage}
%%
%\begin{minipage}[b]{0.5\linewidth}
%  \centering \includegraphics*[width=0.85\columnwidth]{images/fig1b.eps}
%\end{minipage}\\
%   \caption{\label{fig:eco} Example of metadata mapping between two version of video. The metadata in this example are English (left top) and French (left bottom) subtitles.
%   Using content-based signature, similar fragments of the two sequences (shown in red) are aligned and French subtitles are made accessible to the user watching the English version of the video (right).}
%%    \end{center}
%\end{figure*}

%In this paper, we study \emph{content-based synchronization and mapping} of metadata across video versions. %, conceptually visualized in Figure~\ref{fig:eco} (right).

A way to reconcile between the timelines of different versions of a video and the associated metadata is by using \emph{content-based synchronization}.
For this purpose, a
time-dependent signature is computed for each video, allowing to match and align similar parts in different versions of the video, thus giving a translation from one system of time coordinates to another.
In a prototype application %(demonstrated in the supplementary material),
consisting of a client and server, the signature is computed in real-time during the video playback on the client side and sent to the server where it is matched to a database of video signatures. After having established the correspondence to a database sequence, the corresponding metadata on the server side is sent to the client.
With this approach, it is sufficient to keep a database of video signatures computed from some prototype sequence with synchronized metadata.
A new version of the video, previously unseen and coming from any source (e.g. read-only media, streaming, etc.), can be matched to the prototype timeline and the corresponding metadata retrieved. Thus, at least theoretically, any video can be enriched with metadata, provided that similar videos have signatures in the database.

%{\bf [FIGURE 2: prototype client-serve application] Will probably move to supplementary material}

The described application poses some requirements on the signature construction and matching algorithms. First, they should be able to handle large amounts (thousands or millions of hours) of data. This, in turn, imposes the requirements that the signature is compact, easily indexable, and can be searched and matched fast. Secondly, the signature computation should be efficient, and ideally computed in real-time. Finally and most importantly, two versions of a video may different significantly due to
post-production processing (e.g. resolution and aspect ratio change, cropping, color and contrast modification, overlay of logos, compression artifacts, blur, etc.)
and editing (e.g. advertisement insertion or adaptation of a movie for a certain rating category). The signature matching algorithm must therefore be able to cope with such modifications.

Surprisingly, similar problems are encountered in an apparently unrelated field of genetic research, where one of the main problems is matching of DNA and protein sequences.
Many recent efforts, including the notorious Gene Bank and Human Genome projects, resulted in having large collections of annotated DNA and protein sequences, in which newly discovered sequences can be looked up.
The problem of post-processing distortions and editing is analogous to \emph{mutations} occurring in biological DNA sequences.
The scale of genetic data is comparable to that of video sequences (for example, the human genome contains sequences with nearly 3 billion symbols \cite{HGP04:Nature}).
Over the past decades, many efficient methods have been developed for the analysis of genomic sequences, giving birth to the field of \emph{bioinformatics} \cite{mount:04}.

In this paper, we borrow well-established bioinformatic methods for the analysis of video, which can be considered similarly to DNA sequences as shown in Section~\ref{sec:vdna}.
A prototype application considered and shown in the supplementary materials is content-based metadata mapping between versions of video.
%
%Secondly,
The central problem in this application is finding correspondences between video sequences.
In Section~\ref{sec:search}, we draw the analogy to genomic research, which allows to employ dynamic programming sequence alignment \cite{NW:70,SWAT:81} and its fast heuristics \cite{FASTA,BLAST:90}, as well as multiple sequence alignment and phylogenetic analysis \cite{MSA:89}.
Exploring the analogy between mutations in genetic sequences and post-production processing and editing in video, we propose in Section~\ref{sec:metric} a generative approach for learning invariance to such mutations by means of metric learning.
%
%Finally, we show the analogy between dynamic programming sequence alignment algorithms and the raster scan fast marching methods for distance computation, and propose an efficient algorithm for continuous-time video alignment.
%
We obtain a very compact representation ($64$ bit per second of video), which is robust to video transformations and allows efficient indexing and search.
Section~\ref{sec:res} presents experimental results demonstrating the robustness and efficiency of the proposed approach in a variety of applications, including video retrieval and alignment in a large-scale (1K hours) database.
Finally, Section~\ref{sec:concl} concludes the paper.

\subsection{Related work}
\label{sec:prior}

The problem of metadata mapping addressed in this paper is intimately related to \emph{content-based copy detection} and search in video \cite{Laptev,schmid1}.
There, one tries to find copies of a video that has undergone modifications (whether intentional or not) that potentially make it very different visually from the original.
This problem should be distinguished from \emph{action and event recognition} \cite{zelnik2006statistical,boiman2007detecting,laptev1learning}, where the similarity criterion is semantic.
Broadly speaking, copy detection problems boil down to \emph{invariant retrieval} (finding a video invariant to a certain class of transformations)
and action recognition are problems of \emph{categorization} (recognizing a certain class of behaviors in video).
To illustrate the difference, imagine three video sequences: a movie quality version of Star Wars, the same version broadcast on TV with ad insertion and captured off screen with a camcorder, and the lighsabre fight scene reenacted by amateur actors.
The purpose of copy detection is to say that the first and the second video sequences are similar; action recognition, on the other hand, should find similarity between the second and third videos.

One of the cornerstone problems in content-based copy detection and search is the creation of a video representation that would allow to compare and match videos across versions.
Different representations based on mosaic \cite{irani1996efficient}, shot boundaries \cite{indyk1999finding}, motion, color, and spatio-temporal intensity distribution \cite{hampapur2002comparison}, color histograms \cite{li2005video}, and ordinal measure \cite{hua2004robust}, were proposed.
When considering large variability of versions due to post-production modifications, methods based on spatial \cite{low:IJCV:04,matas2004rwb,bay2006ssu} and spatio-temporal \cite{laptev2005space} points of interest and local descriptors were shown to be advantageous \cite{joly2004feature}.
In addition, these methods proved to be very efficient in image search in very large databases \cite{siv:zis:CVPR:03,chum2007scalable}.
More recently, Willems \emph{et al.} \cite{willems2008spatio} proposed feature-based spatio-temporal video descriptors combining both visual information of single video frames as well as the temporal relations between subsequent frames.

One of the main disadvantages of existing video representations is a \emph{constructive approach} to invariance to video transformations.
Usually, the representation is designed based on quantities and properties of video insensitive to typical transformations.
For example, using gradient-based descriptors \cite{low:IJCV:04,bay2006ssu} is known to be insensitive to illumination and color changes.
Such a construction may often be unable to generalize to other classes of transformations, or result in a suboptimal tradeoff between invariance and discriminativity.

An alternative approach, adopted in this paper, is to \emph{learn} the invariance from examples of video transformations. By simulating the post-production and editing process, we are able to produce pairs of video sequences that are supposed to be similar (different up to a transformations) and pairs of sequences from different videos supposed to be dissimilar.
Such pairs are used as a training set for similarity preserving hashing and metric learning algorithms \cite{Shakh:PhD,jai:kul:gra:CVPR:08,torralba1} in order to create a metric between video sequences that achieves optimal invariance and discriminativity on the training set.

%\section{Background}
%\label{sec:backgr}
%
%Since the first full sequencing of a bacteriophage genome done in 1977 completely by hand, DNA sequences of hundreds of different species have been decoded.
%Compared to the tiny phage, the human genome analysis completed three decades later, required completely new computational methodologies.
%The growing amount of genetic data has driven the emergence of \emph{bioinformatics}  %, a field of computer science using string matching and search algorithms
%for automatic analysis of DNA sequences.
%%
%We exploit the analogy between genetic sequences and video, which has recently experienced a similar rapid growth in available data size, and propose an approach to video analysis motivated by genomic research. Representing video information as \emph{video DNA} sequences and applying bioinformatic algorithms allows to search, match, and compare videos on Internet scale.
%

\section{Video DNA}
\label{sec:vdna}

\begin{figure}[th]
%  \centering \includegraphics*[width=1\columnwidth]{images/features.eps}
  \centering \includegraphics*[width=1\columnwidth,bb=1 1 1470 1320]{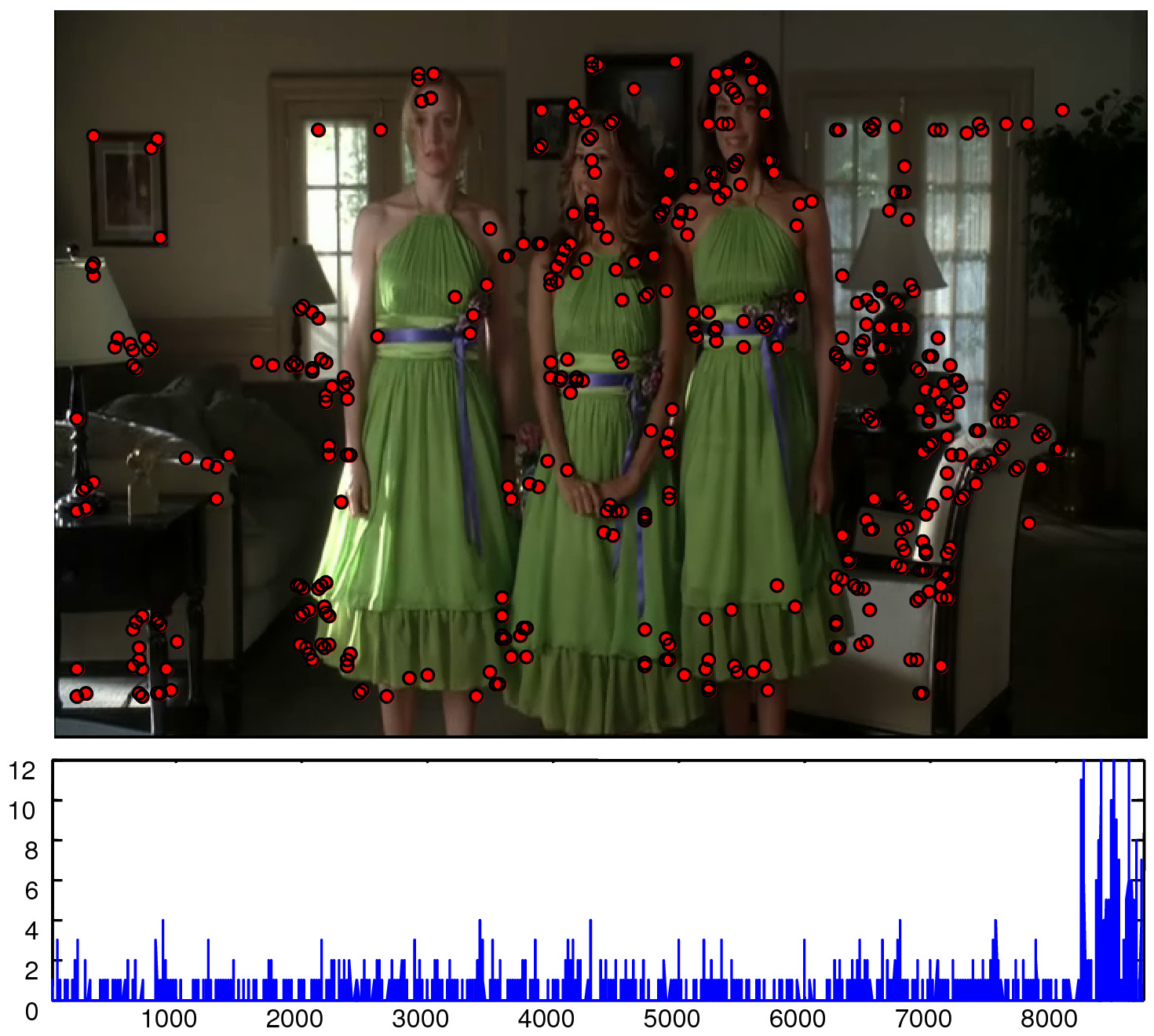}
   \caption{\label{fig:descriptor} {Construction of the visual nucleotides. Top: features detected in a video frame; bottom: corresponding bag of features. After applying similarity-preserving hashing to the bag of feature, the frame is represented by the 64-bit binary word 223E9DF01ADB3E00.}}
\end{figure}

Biological DNA data encountered in bioinformatic applications are long sequences consisting of four letters (representing aminoacids in the DNA molecule, denoted as A, T, C, G and referred to as \emph{nucleotides}).
Extending this example to our problem, one can conceptually think of video as of a sequence of visual information units, which can be represented over some potentially very large alphabet of visual concepts, resulting in a sequence of ``letters'' (or \emph{visual nucleotides}) which we call \emph{video DNA} by analogy to genetic sequences.
Video DNA sequencing, the process of creating a video DNA sequence out of a video, is performed by computing descriptors for each frame (or short sequence of frames) and arranging them on the video timeline (see Figure~\ref{fig:descriptor}).

In this paper, we used a feature-based representation following the standard bag of features paradigm \cite{siv:zis:CVPR:03,chum2007scalable}.
For each frame in the video, we scale down to horizonal resolution of $320$, detect feature points, and compute local image descriptors around these
points using a modification of the \emph{speeded-up robust features} (SURF) \cite{bay2006ssu} feature detection and description algorithm (Figure~\ref{fig:descriptor}, top).
$450$ strongest feature points are used.
Each feature point is described by a $64$-dimensional grayscale and $16$-dimensional color descriptor.
Second, the local descriptors are quantized using the $k$-means clustering algorithm, separately for the grayscale and color feature descriptors, creating grayscale and color \emph{visual vocabularies}.
Vocabulary of $2048$ and $124$ visual words are used for grayscale and color descriptors, respectively.
Each local feature descriptor is replaced by the index of the nearest visual word in the vocabulary.
Third, each frame is divided into four quadrants with 10\% overlap and a \emph{bag of features} (histogram of visual words) in each quadrant is computed.
Four concatenated histograms yield a vector of size $d=8688$ which is used as the frame descriptor (Figure~\ref{fig:descriptor}, bottom).
Fourth, a median of frame descriptors in fixed time intervals is computed, creating the video DNA sequence.
The intervals taken are of size $T$ with step $\Delta_T$. A typical choice is $T=2 sec$ and $\Delta_T = 1 sec$.
%We use intervals of 2 \emph{sec} with 50\% overlap and step of 1 \emph{sec}.

The resulting video DNA is a timed sequence of $d$-dimensional bags of features, which we call \emph{visual nucleotides} by analogy to biological DNA sequences.
The similarity of two video sequences can be quantified by measuring the distance between the corresponding visual nucleotides, which we denote here by $d_\mathcal{A}$. In the simplest case, a Euclidean distance in $\mathbb{R}^d$ is used. In \cite{siv:zis:CVPR:03}, it was shown that a Euclidean distance weighted by the statistical distribution of visual words (\emph{term frequency-inverse document frequency} or tf-idf) is a better way to compare bags of features. We will address the construction of an optimal distance between visual nucleotides in Section~\ref{sec:metric}.

%[FIGURE 3: descriptor construction]

\section{Search and alignment}
\label{sec:search}

%\subsection{Video alignment}

%
Dynamic programming methods used to align biological DNA sequences, notably, the Needleman-Wunsch (NW) \cite{NW:70} and Smith-Waterman (SWAT) \cite{SWAT:81} algorithms, can be applied to finding correspondence between versions of video sequences.

Let $\mathbf{x} = (x_1,...,x_M)$ and $\mathbf{y}=(y_1,...,y_N); x_i, y_i \in \mathbb{R}^d$ %\mathcal{A}$
be two video DNA sequences representing two versions of a video obtained by temporal editing. % (here, $\mathcal{A}$ denotes the alphabet of visual nucleotides).
In this case, $\mathbf{x}$ and $\mathbf{y}$ will typically have locally similar sequences of nucleotides.
In order to find such similarities, we look for an optimal \emph{local alignment} between $\mathbf{x}$ and $\mathbf{y}$, i.e., such a correspondence of indices $\{1,...N\}$ and $\{1,...,M\}$ that on one hand will make the corresponding nucleotides the most similar and on the other will contain gaps of minimum total length. The quality of the correspondence is represented by a similarity score, taking into consideration both the similarity of the nucleotides and the gaps.

The minimum dissimilarity score between the substring of $\mathbf{x}$ of length $i$ and substring of $\mathbf{y}$ of length $j$ is given by the following recursive equation,
\begin{eqnarray}
\label{eq:swat}
s_{ij} &=& \max \left\{
\begin{array}{cc}
  0 &  \\
  s_{i-1,j-1} + d(a_i,b_j) & \mathrm{match} \\
  s_{i-1,j} + g(a_i)& \mathrm{deletion} \\
  s_{i,j-1} + g(b_j)& \mathrm{insertion}
\end{array}
\right\},
\end{eqnarray}
where $i=1,...,M, j=1,...,N$ and $s_{i0} = s_{0j}=0$ for all $i=0,...,M, j=0,...,N$.
$d_\mathcal{A}(a,b)$ is the similarity between nucleotides and $g(a)$ is the \emph{gap penalty}.
The values of $s$ are determined by means of dynamic programming and the optimal correspondence is established by backtracking \cite{SWAT:81}.

\subsection{Fast heuristics}

The main disadvantage of dynamics programming alignment methods is their high complexity of $\mathcal{O}(NM)$. In our application, when a short sequence ($N$ of order of $10^3-10^4$ for a typical movie assuming $\Delta_T = 1sec$) is compared to a large database containing signature of thousands or millions of hours of videos ($M$ in the order of $10^6-10^9$), such an approach may be computationally prohibitive.
A similar complexity problem is encountered in gene search applications in bioinformatics, where typical databases contain sequences totalling in  millions or billions of letters.

To overcome this problem, fast heuristics such as FASTA \cite{FASTA} and BLAST \cite{BLAST:90} have been developed.
The key idea of these approaches is to first locate matches of short combinations of nucleotides of fixed size $k$ (typically ranging between 2 and 10), which establish multiple coarse initial correspondence between regions in the two sequences.
Using search engine terminology, the initial correspondence established by FASTA/BLAST algorithms are a \emph{short list} of candidates.
The correspondence is later refined using a banded version of the SWAT algorithm, applied on sequences around the initial regions.
At this stage, video DNA sequences at higher temporal resolution can be used.

%{\bf ADD: multiple resolutions}

%
%\subsection{Continuous time alignment}
%
%Multiple resolutions
%
%Problem of sequences alignment: dependent on sampling, problem with time scaling
%Analogy to metrication errors
%
%Relation to raster scan FMM and continuous time alignment
%
%
%[FIGURE: metrication errors, continuous time alignment]

\subsection{Multiple sequence alignment}

In many cases, it is desired to find alignment between more than two videos, a problem analogous to \emph{multiple sequence alignment} (MSA) in bioinformatics. MSA is used in phylogenetic analysis \cite{MSA:89}, in order to discover evolutionary relations between DNA sequences.
In video, a similar problem is \emph{version control}, where multiple versions of a video are given and one wishes to establish, for example, from which source they were derived and which sequence was the original.

Straightforward generalization of dynamic programming alignment algorithms to MSA results in an exponential complexity.
For this reason, sub-optimal heuristics such as progressive sequence alignment are used.
For example, in CLUSTAL \cite{CLUSTAL}, first all pairs of sequences are aligned separately. Alignment cost acts as a measure of the pair-wise sequence dissimilarity.
Given the pairwise dissimilarity matrix, a \emph{guide tree} is constructed by means of clustering (e.g. neighorhood joining).
Finally, series of pair-wise alignments following the branching order in the tree are performed. This way, most similar sequences are aligned first and most dissimilar last (for detailed algorithm description, see \cite{CLUSTAL}).

\section{Mutation-invariant metric}
\label{sec:metric}

Post-production transformations in video are analogous to \emph{mutations} in biological DNA sequences and can be manifested either as insertion or deletion of visual nucleotides (\emph{indel mutations}) as a result of temporal editing, or as \emph{substitution mutations}, in which the visual content is replaced by another as the result of spatial editing such as resolution or aspect ratio change, cropping, compression artifacts, overlay of subtitles or channel logo, etc.
While local alignment is efficient in coping with insertion or deletion mutations by proper selection of the gap penalty, substitution mutations can be a major challenge, as they may have a global effect on the entire video DNA sequence (imagine, for example, that due to non-uniform scaling of the video, the bag of features changes in every frame).

In biological DNA sequence analysis, the exact mechanism of mutations is not completely understood or reproduced; therefore, empirical models of nucleotide mutation probability are used \cite{PAM:78}. In our case, on the other hand, it is easy to reproduce the post-production processing that causes mutation in video DNA.
Ideally, our visual nucleotides should be discriminative (such that two intervals belonging to different videos are dissimilar)
and invariant (such that two transformations of the same interval are similar).
Though our construction of visual nucleotides rely on feature descriptors that are insensitive to certain transformations of the frame (scale, mild brightness and contrast variations), other transformations (e.g. cropping, subtitle overlay, etc.)
may result in different visual nucleotides. As a consequence, the simple Euclidean metric would not be invariant under such transformations.

Yet, it is possible to \emph{learn} the best mutation-invariant metric between nucleotides on a training set.
Assume that we are given a set of nucleotides $\XX$ describing different intervals of video, and $\TT$ the class of all transformations invariance to which is desired.
We denote by $\PP = \{ (x, x\circ \tau) : x \in \XX, \tau \in \TT \}$ the set of all \emph{positive pairs} (visual nucleotides of identical intervals, differing up to some transformation), and by $\NN \subset \XX \times \XX$ the set of all \emph{negative pairs} (visual nucleotides of different intervals).
Negative pairs are modeled by sampling numerous intervals from different videos, which are known to be distinct. For positive pairs, we generate representative transformations from class $\TT$.
Our goal is to find a metric between nucleotides that ideally is as small as possible on the set of positives and as large as possible on the set of negatives.
%
%Alternatively, we can fix a metric space and find a projection from the space of visual nucleotides ($\mathbb{R}^d$) to this space.

Shakhnarovich \cite{Shakh:PhD} considered metric parameterized as
\begin{eqnarray}
d_{A,b}(x,x') &=& d_\mathbb{H}(\mathrm{sign}(Ax + b), \mathrm{sign}(Ax' + b)),
\end{eqnarray}
where
\begin{eqnarray}
d_\mathbb{H}(\xi,\xi') &=& \frac{n}{2} - \frac{1}{2} \sum_{i=1}^n \mathrm{sign}(\xi_i \xi'_i),
\end{eqnarray}
is the \emph{Hamming metric} in the $n$-dimensional \emph{Hamming space} $\mathbb{H}^n = \{-1,+1\}^n$ of binary sequences of length $n$.
$A$ and $b$ are an $n \times d$ matrix and an $n \times 1$ vector, respectively, parameterizing the metric. Our goal is to find $A$ and $b$ such that $d_{A,b}$ reflects the desired similarity of pairs of visual nucleotides $x,x'$ in the training set.

%Following \cite{Shakh:PhD}, we use the $n$-dimensional \emph{Hamming space} $\mathbb{H}^n$ of binary sequences of length $n$, with the \emph{Hamming metric}
%\begin{eqnarray}
%d_\mathbb{H}(\xi,\xi') &=& \frac{n}{2} - \frac{1}{2} \sum_{i=1}^n \mathrm{sign}(\xi_i \xi'_i),
%\end{eqnarray}
%%
%where $\xi_i \in \{-1,1\}, \, i = 1,...,n$.
%%
%%
%We are looking for a projection of the form
%\begin{eqnarray}
%\xi(x) &=& \mathrm{sign}(Ax + b),
%\end{eqnarray}
%with $A$ and $b$ being an $n \times d$ matrix and an $n \times 1$ vector, respectively, and try to find $A$ and $b$ such that $d_\mathbb{H}(Ax+b,Ax'+b)$ reflects the desired similarity of pairs of visual nucleotides $x,x'$ in the training set.

%
%\begin{figure}[th]
%  \centering \includegraphics*[width=1\columnwidth]{images/hashing.eps}
%   \caption{\label{fig:hashing} Similarity-preserving hashing aims at finding a projection $\xi : \mathrm{R}^d \rightarrow \mathbb{H}^n$ from the visual nucleotide space to the $n$-dimensional Hamming space rendering positive pairs ($x,\tau x$) similar and negative pairs ($x,y$ and $\tau x, y$) dissimilar (top). The optimal projection is found by minimizing the false positive and false negative rates (bottom). }
%\end{figure}
%

Ideally, we would like to achieve $d_{A,b}(x,x') \le d_0$ for $(x,x') \in \PP$, and $d_{A,b}(x,x') > d_0$ for $(x,x') \in \NN$, where $d_0$ is some threshold.
In practice, this is rarely achievable as the distributions of $d_{A,b}$ on $\PP$ and $\NN$ have cross-talks responsible for false positives ($d_{A,b} \le d_0$ on $\NN$) and false negatives ($d_{A,b} > d_0$ on $\PP$).
%However, we can aim at minimizing such cross-talks, e.g., by minimizing the exponential loss
%
Thus, optimal $A,b$ should minimize these cross-talks,
\begin{eqnarray}
\label{eq:exp-loss}
\min_{A,b} && \frac{1}{|\PP|} \sum_{(x,x') \in \PP} \left\{e^{\mathrm{sign}(d_{A,b}(x,x') - d_0 )}\right\} + \nonumber\\
           &&
           \frac{1}{|\NN|} \sum_{(x,x') \in \NN} \left\{e^{\mathrm{sign}(d_0 - d_{A,b}(x,x')  )}\right\}.
\end{eqnarray}
%
%Though the distributions of descriptors in $\PP$ and $\NN$ are very complicated and difficult to model, the expectations on $\PP$ and $\NN$ can be replaced by averages.

In \cite{Shakh:PhD}, Shakhnarovich proposed considering learning optimal parameters $A,b$ as a boosted binary classification problem, where $d_{A,b}$ acts as a strong binary classifier, and each dimension of the linear projection $\mathrm{sign}(A_k x + b_k)$ can be considered as a weak classifier. This way, AdaBoost algorithm can be used to progressively construct $A$ and $b$,
which would be a greedy solution of (\ref{eq:exp-loss}).
At the $k$-th iteration, the $k$-th row of the matrix $A$ and the $k$-th element of the vector $b$ are found minimizing a weighted version of (\ref{eq:exp-loss}). Weights of false positive and false negative pairs are increased, and weights of true positive and true negative pairs are decreased, using the standard Adaboost reweighting scheme \cite{Adaboost}.
%
%
%Optimal $A_k$ should satisfy
While it is difficult to find $A_k$ minimizing~(\ref{eq:exp-loss}) because of the non-linearity, we found that the minimizer of the exponential loss is related to another simpler problem,
\begin{eqnarray}
\label{eq:opt-dir}
A_k %&=& \mathrm{arg}\max_{\|A_k\| = 1}\, \frac{\displaystyle{ \mathop{\mathrm{E}}_{(x,x') \in \NN} (A_k^\Tr x - A_k^\Tr x')^2  } }{\displaystyle{ \mathop{\mathrm{E}}_{(x,x') \in \PP} (A_k^\Tr x - A_k^\Tr x')^2  } } \nonumber\\
%&=& \mathrm{arg}\max_{\|a\| = 1}\, \frac{\displaystyle{ a^\Tr \mathop{\mathrm{E}}_{(x,x') \in N} (x-x')(x-x')^\Tr \, a  } }{\displaystyle{ a^\Tr \mathop{\mathrm{E}}_{(x,x') \in P} (x-x')(x-x')^\Tr \, a  } } \nonumber\\
&=& \mathrm{arg}\max_{\|A_k\| = 1}\, \frac{A_k^\Tr C_{\NN} A_k}{A_k^\Tr C_{\PP} A_k},
\end{eqnarray}
%where the expectations are taken with respect to the current sample weights, and
where $C_{\PP}$ and $C_{\NN}$ are the covariance matrices of the positive and negative pairs, respectively. It can be shown that $A_k$ maximizing (\ref{eq:opt-dir}) is the largest generalized eigenvector of $C_{\NN}^\frac{1}{2} A_k = \lambda_\mathrm{max} C_{\PP}^\frac{1}{2} A_k$.
%
%This approach is similar in its spirit to \emph{linear discriminative analysis} (LDA).
%
Since the minimizers of~(\ref{eq:exp-loss}) and~(\ref{eq:opt-dir}) do not coincide exactly, in our implementation, we select a subspace spanned by the largest ten eigenvectors, out of which the direction as well as the threshold parameter $b$ minimizing the exponential loss are selected.

There are a few advantages to the described approach.
First, the metric $d_{A,b}$ is constructed to achieve the best discriminativity and invariance on the training set. If the training set is sufficiently representative, such a metric generalizes well.
It can be used as $d_\mathcal{A}$ in the alignment and search algorithms described in Section~\ref{sec:search}.
Secondly, the projection itself has an effect of dimensionality reduction, and results in a very compact representation of visual nucleotides as
\emph{bitcodes} (for example, the frame shown in Figure~\ref{fig:descriptor} is represented by the hexadecimal word 223E9DF01ADB3E00).
Such bitcodes can be efficiently stored and manipulated in standard databases.
Thirdly, modern CPU architectures allow very efficient computation of Hamming distances using bit counting and SIMD instructions. Since each of the bits can be computed independently, score computation in the alignment algorithm can be further parallelized on multiple CPUs using either shared or distributed memories. Due to the compactness of the bitcode representation, search can be performed in memory (a single 8GB memory system is sufficient to store about 300,000 hours of video with $1$ second resolution for $n=64$).

\section{Results}
\label{sec:res}

%Overview of experiments.

In the experimental validation, we worked with a database containing 1013 hours of assorted video content (movies, 2D and 3D cartoons, talk shows, sports) taken from DVDs.
Video DNA sequences were computed with parameters $T=2sec$ and $\Delta_T = 1sec$.
Hamming space of dimension $n=64$ was used for bitcode representation.
Metric learning was performed offline on a training set containing $2\times 10^5$ positive and $8\times 10^5$ negative pairs. Positives were created using
transformation simulated with AviSynth frame server.  %had five strengths: levels 1-3 typically encountered in video editing, and levels 4-5 representing extreme transformations, used to test when the algorithms fail (Figure~\ref{fig:xform_strength}).

\begin{figure*}[tp]
    \begin{center}
    \includegraphics[width=1\linewidth,bb=1 1 1440 390]{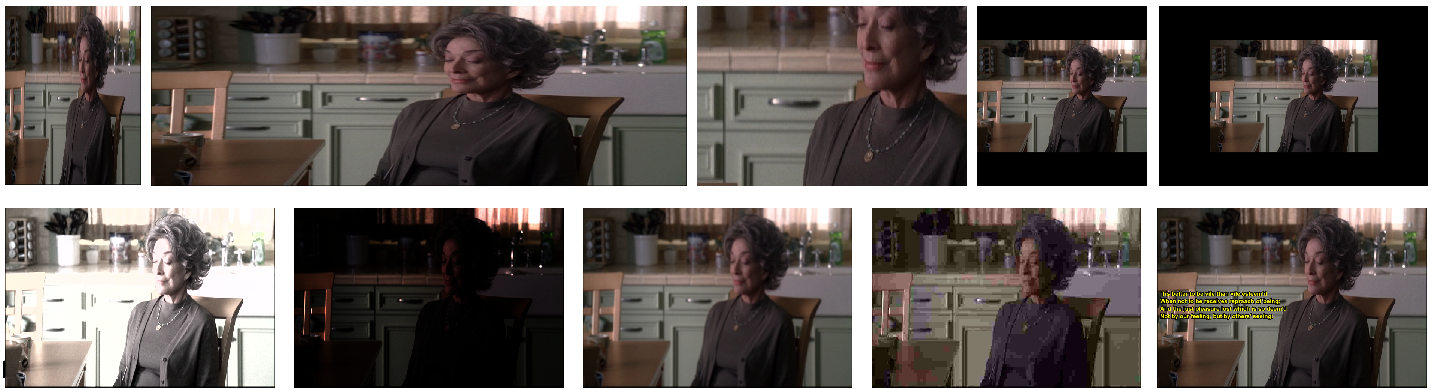}
   \caption{\label{fig:xforms} \small Examples of transformations used in our experiments. Top: geometric transformations (non-uniform scale, cropping, letter box and borders). Bottom: pixel transformations (gamma, blur, quantization, subtitles).}
   %different transformation classes used in the experiments, left to right top down: blur, soften, sharpen, brighten, darken, contrast low, contrast high, saturate, desaturate, quantization, overlay, vertical and horizontal crop, horizontal crop, vertical crop, letterbox, pillarbox, border, horizontal non-uniform scale, vertical non-uniform scale, uniform scale. Bottom: different strengths of the  border transformation.}
    \end{center}
\end{figure*}

%\subsection
\mypara{Large scale search}
For the evaluation of search and alignment, we used a scheme proposed by \cite{Laptev}. Randomly selected short sequences from the database were used as queries. The queries were constructed in such a way that there was exactly one correct match with the database. In BLAST and FASTA-type algorithms, the queries represent the short nucleotide sequences used to establish initial matches. The queries underwent transformations (shown in Figure~\ref{fig:xforms})
typical for the video post-production, including spatial and pixel transformations (cropping, letter and pillar box, contrast and color balance,
compression noise, resolution and aspect ratio change, subtitle overlay) and temporal transformations (framerate change and time shift).
Each transformation appeared at multiple strengths (denoted as 1--3).

%
%
%
%\begin{figure*}[tp]
%    \begin{center}
%    \includegraphics[width=0.95\linewidth]{images/xform_strength.eps}
%   \caption{\label{fig:xform_strength} \small Different strength (shown left to right 1 to 5) of the Darken transformation. }
%    \end{center}
%\end{figure*}

Short sequences locally matching to the queries were found in the database using a FASTA/BLAST-type algorithm described in Section~\ref{sec:search}.
The matching precision was measured as precision with recall of $1$, i.e., the percentage of correct first matches.
Matches were considered correct if they were within 1 \emph{sec} tolerance off the groundtruth match (i.e., falling within the temporal resolution of our representation).
%
%
%Examples of correct and incorrect matches are presented in Figure~\ref{fig:matches}. Figure~\ref{fig:top10} depicts the short list of ten top matches found for a query sequence.
%
Typical search time was $250$ \emph{msec}.

Table~\ref{tab:perf:breakdown} shows the breakdown of search precision according to transformations types and strengths. 10,770 queries of 10 \emph{sec} length were used.
Table~\ref{tab:perf:length} shows the search precision as function of the query length (varying from 5 to 30 \emph{sec}), on a query set of 20,160 queries, including all transformations of strength 1--3. It shows that 10 \emph{sec} of video are sufficient to achieve less than 3\% search error in a database of 1013 hours across versions including significant transformations. This number falls below 1\% for a 20 \emph{sec} query.

\begin{figure*}[ht!]
    \begin{center}
    \includegraphics[width=1\linewidth,bb=1 1 1780 595]{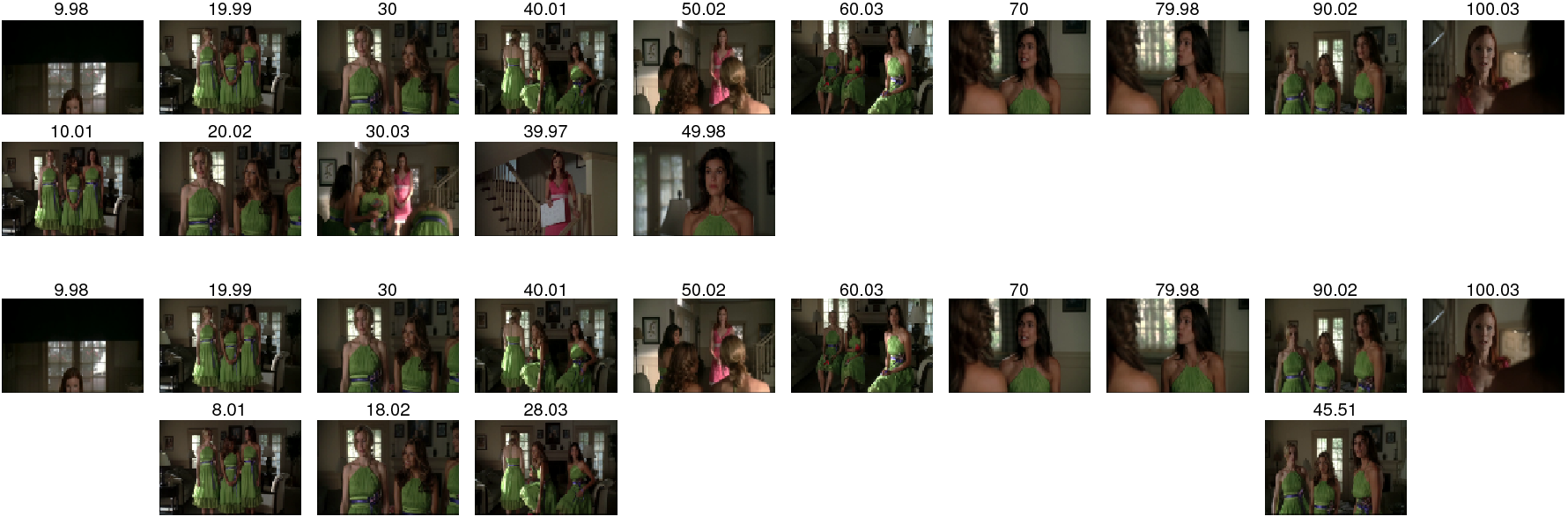}
   \caption{\label{fig:align} \small Top: two versions of a video have different timelines because of editing. Bottom: alignment based on Video DNA brings the two timelines in correspondence.}
   %different transformation classes used in the experiments, left to right top down: blur, soften, sharpen, brighten, darken, contrast low, contrast high, saturate, desaturate, quantization, overlay, vertical and horizontal crop, horizontal crop, vertical crop, letterbox, pillarbox, border, horizontal non-uniform scale, vertical non-uniform scale, uniform scale. Bottom: different strengths of the  border transformation.}
    \end{center}
\end{figure*}

%\subsection{Local alignment}
\mypara{Local alignment}
In order to evaluate the performance of local alignment, we performed alignment of sequences from subset of the database containing approximately 300 hours of video using the dynamic programming algorithm described in Section~\ref{sec:search}. %Smith-Waterman algorithm.
Query sequences underwent spatial transformations from the previous experiments, as well as different temporal transformations. The latter included deletion of portions of video, substitution with other videos, and insertion of blackness periods (both with sharp or gradual fade-in and fade-out transitions of different durations); local speeding up and slowing down of the video playback speed; and removal of significant parts of the original footage from the query sequence.
Table~\ref{tab:perf:breakdown:temporal}  shows the breakdown of alignment precision according to transformations types and strengths.
%$1$ \emph{sec} resolution video DNA sequences were used.
%
An example of two aligned versions of a sequence from the \emph{Desperate Housewives} series is shown in Figure~\ref{fig:align}.

\begin{figure}[tp]
    \begin{center}
    \includegraphics[width=1\linewidth,bb=1 1 1441 800]{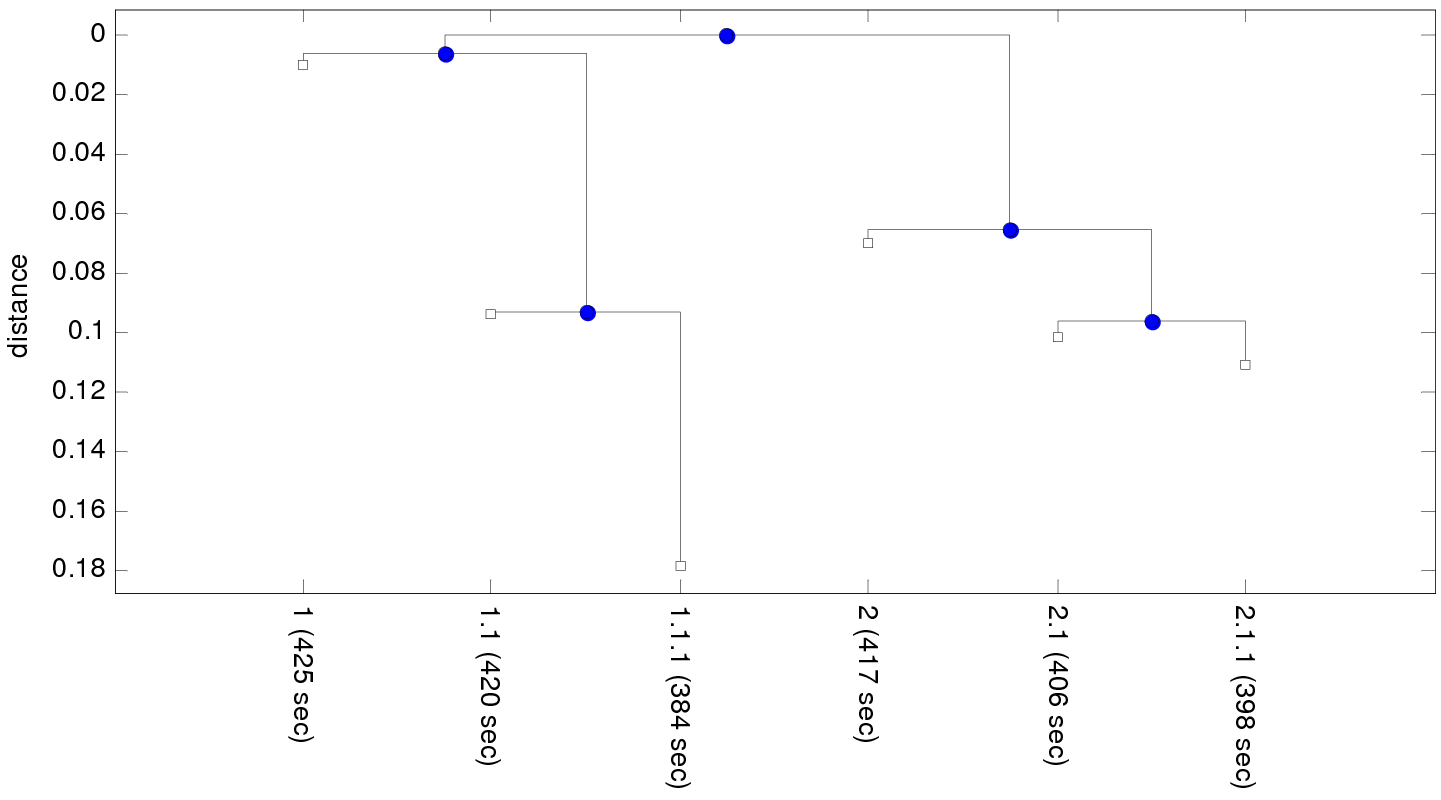}
   \caption{\label{fig:phylo} \small Phylogenetic analysis of six versions of \emph{Desperate Housewives}. Each leave in the dendrogram represents a sequence, labeled according to its version (e.g., 1.1. means the sequence was derived from sequence 1 by means of removing a shot).
   Vertical axis represents the distance between the versions, computed as the percentage of dissimilar parts (gaps). Evolutionary relations between
   versions can be clearly inferred from the dendrogram.    }
   %different transformation classes used in the experiments, left to right top down: blur, soften, sharpen, brighten, darken, contrast low, contrast high, saturate, desaturate, quantization, overlay, vertical and horizontal crop, horizontal crop, vertical crop, letterbox, pillarbox, border, horizontal non-uniform scale, vertical non-uniform scale, uniform scale. Bottom: different strengths of the  border transformation.}
    \end{center}
\end{figure}

\mypara{Phylogenetic analysis}
Figure~\ref{fig:phylo} shows a dendrogram representing the evolutionary relations between six versions derived from the \emph{Desperate Housewives} from Figure~\ref{fig:align}.
Version x.y was obtained by removing a shot from sequence x.
The dendrogram was constructed from the matrix of pairwise sequence distances (computed as the ratio of the gaps length to the total sequence length in aligned pairs of sequences) using neighbor joining approach.
One can clearly see how subsequent versions were derived.

\begin{table}[t!]
   \caption{ \small Precision (percentage of first matches falling within $1$ \emph{sec} tolerance) broken down according to transformation types and strengths. }
\begin{center}
\begin{tabular}{lccc}
\hline\hline
  &  &  \small Strength  &    \\
  \cline{2-4}
\small Transform. &  \small 1  & \small 2 & \small 3   \\
\cline{1-4}

   \small Blur         & \small  100.00 & \small  100.00 & \small  100.00  \\
   \small Soften       & \small  100.00 & \small  98.81  & \small  98.81   \\
   \small Sharpen      & \small  100.00 & \small  100.00 & \small  100.00  \\
   \small Brighten  & \small  100.00 & \small  100.00 & \small  99.21   \\
   \small Darken    & \small  100.00 & \small  99.60  & \small  95.63   \\
   \small Contrast  & \small  99.80  & \small  100.00 & \small  98.21   \\

   \small Saturation    & \small  100.00 & \small  98.81  & \small  91.47   \\

   \small Quantization & \small  88.89  & \small  86.90  & \small  90.08   \\

   \small Overlay              &  \small 100.00   & \small 99.91   & \small  98.41       \\

 \hline

   \small Crop              & \small    98.77 & \small    95.59 & \small    89.95 \\
   \small Letterbox         & \small    99.12 & \small    98.59 & \small    97.53 \\
%   \small Rotate            & \small    96.83 & \small    91.53 & \small    49.21  \\
    \small Nonunif. scale  & \small    98.41 & \small    99.21 & \small    96.56 \\

   \small Uniform scale     & \small    100.00 & \small   100.00 & \small   100.00 \\

 \hline

   \small Framerate           &  \small 100.00   & \small 100.00   & \small  100.00      \\

   \small Time shift          &  \small 100.00   & \small 99.91   & \small  99.63      \\

 \hline

%   \small Crop                   &  \small 98.77   & \small 95.59   & \small  89.95      & \small  76.37        & \small  61.02\\
%   \small Letterbox              &  \small 99.12   & \small 98.59   & \small  97.53      & \small  97.00        & \small  96.30\\
%   \small Unif. scale          &  \small 100   & \small 100   & \small  100      & \small  100        & \small  99.47\\
%   \small Non-unif. scale      &  \small 98.41   & \small 99.21   & \small  96.56      & \small  94.18        & \small  90.21\\
% \hline
%   \small All                    &  \small 98.85   & \small 97.5   & \small  91.34      & \small  84.61        & \small  69.52\\
 \hline
\end{tabular}
\end{center}
\label{tab:perf:breakdown}
\end{table}

\begin{table}[t!]
   \caption{ \small Precision (percentage of first matches falling within $1$ \emph{sec} tolerance) as function of query length in seconds. }
\begin{center}
\begin{tabular}{ccccc}
\hline\hline
% &  & \small Length (seq)  &  &   \\
\small \small 5 \emph{sec}  & \small 10 \emph{sec} & \small 15 \emph{sec} & \small 20 \emph{sec} & \small 30 \emph{sec}  \\
 \hline
 \small 93.7   & \small 97.08   & \small  98.36      & \small  99.22        & \small  99.38\\
 \hline
\end{tabular}
\end{center}
\label{tab:perf:length}
\end{table}

\begin{table}[t!]
   \caption{ \small Precision (percentage of first matches within $1$ \emph{sec} tolerance) broken down according to transformation types and strengths. }
\begin{center}
\begin{tabular}{lccc}
\hline\hline
  &  &  \small Strength  &    \\
  \cline{2-4}
\small Transformation &  \small 1  & \small 2 & \small 3    \\
%\cline{1-4}
\hline

   \small Deletion \& insertion     & \small  99.99 & \small    99.96 & \small    99.81     \\
   \small Partial                   & \small  99.96 & \small    99.90 & \small    99.34     \\
   \small Fade ins \& outs          & \small  99.92 & \small    99.16 & \small    99.40     \\
   \small Local speed changes       & \small  99.91 & \small    99.90 & \small    99.90     \\
   \small Substitutions             & \small  98.45 & \small    95.67 & \small    91.01     \\
   \small Overlay                   & \small  99.85 & \small    99.62 & \small    99.28     \\
 \hline \hline
\end{tabular}
\end{center}
\label{tab:perf:breakdown:temporal}
\end{table}

%\subsection{Metadata mapping}

%show how annotation is transferred (YouTube sequences)

%\subsection{Phylogenetic analysis of videos}

%\subsection{Multiple sequence alignment}

%show dendrograms

%------------------------------------------------------------------------
\section{Conclusions}
\label{sec:concl}

We presented a framework for the construction of robust and compact video representations.
By appealing to the analogy between genetic sequences and video, we employed bioinformatics algorithms that allow
efficient search and alignment of video sequences. Also, we showed that using metric learning, it is possible to
design an optimal metric on a training set of generated video transformations.
%
%In future work, we intend exploring bioinformatic methods. For example, taking into consideration the context

We believe that harvesting video and related metadata available in the public domain and creating a database of
annotated video DNA sequences together with search and alignment tools could eventually have an impact similar to
that of the Human Genome project in genomic research.
Having, for example, a large database containing signatures of the most popular Hollywood movies would allow identifying and
synchronizing any version of a movie no matter when, where, and from which source it is played.
The database can be used for finding copies and versions of movies on the web, in order to cope with piracy, enhance video content
with metadata such as subtitles, or provide keywords for contextual advertisement engines.
%
% or clickable objects.
%
Finally, human annotations and semantic information would enable video understanding by using matching annotations of similar
videos from the database. %Annotations of objects in video can be used for adding interactive features or contextual advertisement to video.

%Size of negatives vs positives

%Wald embedding and fast rejection

%Multi-modal hashing

{
\bibliographystyle{plain}
\bibliography{biblio_videodna}
}

\end{document}